\documentclass[10pt,twocolumn,letterpaper]{article}

\usepackage[pagenumbers]{iccv} 

\usepackage{lipsum}

\usepackage{graphicx}
\usepackage{amsmath}
\usepackage{amssymb}
\usepackage{booktabs}
\usepackage{multirow}
\usepackage{stfloats}
\usepackage{caption}
\usepackage{colortbl}
\usepackage{multirow}
\usepackage{multicol}
 \usepackage{subcaption} 
 \usepackage{pifont}
\usepackage{marvosym}
\usepackage{makecell}
\definecolor{linecolor}{RGB}{220,236,211}
\usepackage{booktabs} 

\definecolor{greenbg}{rgb}{0.9, 1.0, 0.9} 

\usepackage{footnote}
\usepackage{url}
\usepackage[flushleft]{threeparttable}
\usepackage{enumitem}
\usepackage{algorithm}
\usepackage{algorithmic}
\usepackage{listings}
\usepackage{tabularx}
\usepackage {pifont} 
\usepackage{bm}
\usepackage{tablefootnote}
\usepackage{balance}
\usepackage{float}
\usepackage{subcaption}
\usepackage[title]{appendix}
\usepackage{marvosym}
\usepackage[accsupp]{axessibility}  
\definecolor{kaiming-green}{RGB}{57,181,74} 

\makeatletter\renewcommand\paragraph{\@startsection{paragraph}{4}{\z@}
  {.5em \@plus1ex \@minus.2ex}{-.5em}{\normalfont\normalsize\bfseries}}\makeatother

\definecolor{iccvblue}{rgb}{0.21,0.49,0.74}
\usepackage[pagebackref,breaklinks,colorlinks,allcolors=iccvblue]{hyperref}

\hypersetup{
    colorlinks=true,
    linkcolor=iccvblue,
    filecolor=magenta,      
    urlcolor=magenta,
    citecolor=iccvblue
}

\def\ourmethod{\textsc{\textbf{Hermes}}}
\def\ours{\textsc{{Hermes}}}

\def\paperID{} 
\def\confName{ICCV}
\def\confYear{2025}

\title{
\ourmethod: A Unified Self-Driving World Model for \\Simultaneous 3D Scene Understanding and Generation }

\author{Xin Zhou$^{1*}$, Dingkang Liang$^{1*\dag}$, Sifan Tu$^{1}$, Xiwu Chen$^{3}$, Yikang Ding$^{2\dag}$, \\ 
Dingyuan Zhang$^{1}$, Feiyang Tan$^{3}$, Hengshuang Zhao$^{4}$, Xiang Bai$^{1\text{\Letter}}$\\
$^{1}$ Huazhong University of Science and Technology, $^{2}$ MEGVII Technology, \\$^{3}$ Mach Drive, $^{4}$ The University of Hong Kong\\
{\tt \{xzhou03, dkliang\}@hust.edu.cn}
}

\begin{document}
\twocolumn[
\maketitle
{
\vspace{-11.5mm}
\begin{figure}[H]
\hsize=\textwidth
\centering
\includegraphics[width=2.1\linewidth]{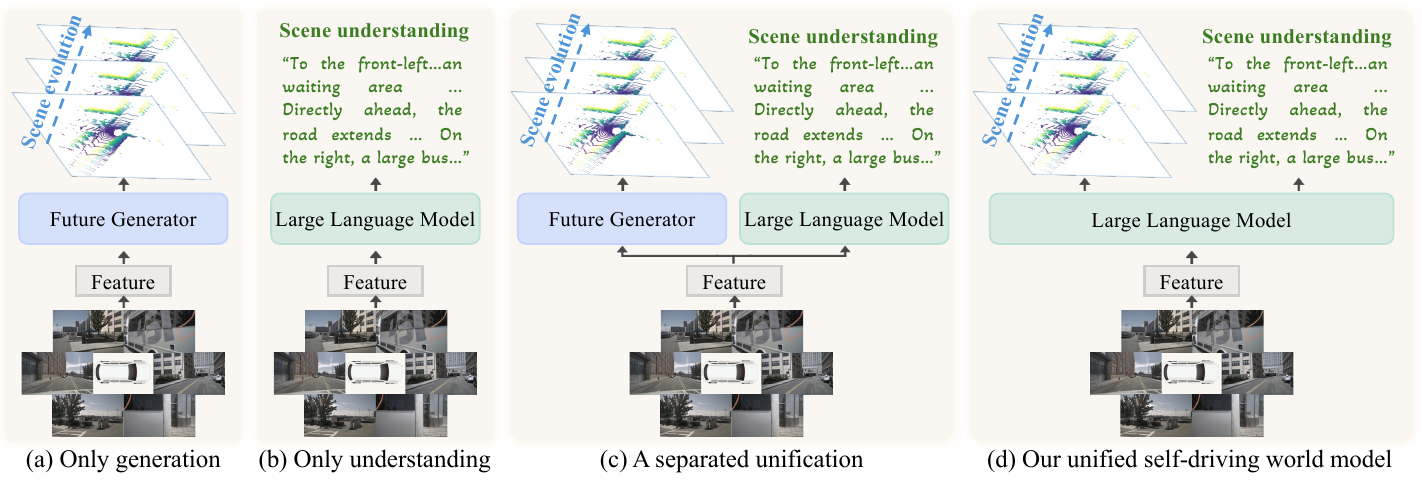}
\vspace{-20pt}
\caption{(a) Previous driving world models focus on generative scene evolution prediction. (b) Large language models for driving are limited to scene understanding. (c) A straightforward unification manner using the generator and large language models separately with a shared feature. (d) The proposed simple framework unifies 3D scene understanding and generates scene evolution based on given actions.}
\label{fig:intro}
\end{figure}
}]

{\let\thefootnote\relax\footnotetext{* Equal contribution. $\dag$ Project leader. $^{\text{\Letter}}$~Corresponding author.}}

\begin{abstract}
Driving World Models (DWMs) have become essential for autonomous driving by enabling future scene prediction. However, existing DWMs are limited to scene generation and fail to incorporate scene understanding, which involves interpreting and reasoning about the driving environment. In this paper, we present a unified Driving World Model named \ourmethod. We seamlessly integrate 3D scene understanding and future scene evolution (generation) through a unified framework in driving scenarios. Specifically, \ours~leverages a Bird's-Eye View (BEV) representation to consolidate multi-view spatial information while preserving geometric relationships and interactions. We also introduce world queries, which incorporate world knowledge into BEV features via causal attention in the Large Language Model, enabling contextual enrichment for understanding and generation tasks. We conduct comprehensive studies on nuScenes and OmniDrive-nuScenes datasets to validate the effectiveness of our method. \ours~achieves state-of-the-art performance, reducing generation error by 32.4\% and improving understanding metrics such as CIDEr by 8.0\%. The model and code will be publicly released at \url{https://github.com/LMD0311/HERMES}.
\end{abstract}

\section{Introduction}
\label{sec:intro}

Driving World Models (DWMs)~\cite{hu2023gaia,gao2024vista,zhao2024drivedreamer,wang2024driving} have become increasingly important in autonomous driving for their ability to predict future scene evolutions. These models simulate potential changes in the surrounding environment, enabling the vehicles to forecast risk, optimize routes, and make timely decisions in dynamic situations. Among the various modalities, point clouds~\cite{hou2023query,li2023dds3d,yang2024visual,liang2025parameter} naturally preserve the geometric relationships between different objects and their surroundings, making them well-suited for accurately describing scene evolutions~\cite{khurana2023point,zyrianov2024lidardm,weng2022s2net,zhang2023learning}.

However, despite the progress in scene generation, a crucial limitation of current DWMs is their inability to incorporate scene understanding fully. Specifically, while these DWMs excel at predicting how the environment will evolve (Fig.~\ref{fig:intro}(a)), they are hard to interpret and describe the environment, answer questions about it, or provide relevant contextual information (i.e., VQA, scene description). 

Recently, vision-language models (VLMs)~\cite{liu2024visual,chen2024internvl,li2024monkey} have achieved remarkable advancements in general vision tasks by leveraging world knowledge and causal reasoning capabilities and have been successfully applied in autonomous driving scenes~\cite{wang2024omnidrive,sima2025drivelm}. As shown in Fig.~\ref{fig:intro}(b), these driving VLMs are capable of performing tasks such as answering complex queries about the driving environment, generating descriptions of scenes, and reasoning about the relationships between various entities. However, while they improve understanding of the current driving environment, they still lack predictive capabilities for how the scene will evolve. This gap limits their effectiveness in autonomous driving, where both 3D scene understanding and future scene prediction are necessary for informed decision-making. This naturally gives rise to the question: \textit{how can world knowledge and future scene evolutions be seamlessly integrated into a unified world model?}

Driven by the above motivation, in this paper, we propose a unified world model that connects both understanding and generation tasks. Our method is referred to as \ourmethod, as illustrated in Fig.~\ref{fig:intro}(d), distinguishes itself from conventional methods, which typically specialize in either generation or scene understanding (e.g., VQA and caption). \ourmethod~extends the capabilities of large language models (LLMs) to simultaneously predict future scenes and understand large-scale spatial environments, particularly those encountered in autonomous driving. However, constructing such a unified model is a highly non-trivial problem, as it requires overcoming several key challenges:

\textbf{Large spatiality in multi-view.} LLMs typically face max token length limits, especially in autonomous driving, where multiple surrounding views must be processed (e.g., six-view images in the nuScenes dataset~\cite{caesar2020nuscenes}). Directly converting these multi-view images into tokens would exceed the token limit and fail to capture the interactions among different views. To address this, we propose to tokenize the input via a Bird's-Eye View (BEV) representation. It offers two key benefits: 1)  BEV effectively compresses the surrounding views into a unified latent space, thus overcoming the token length limitation while retaining key spatial information. 2) BEV preserves geometric spatial relationships between views, allowing the model to capture interactions between objects and agents across multiple perspectives.

\textbf{The integration between understanding and generation.} A straightforward way to unify scene understanding and generation would be to share the BEV features and apply separate models for understanding (via LLMs) and generation (via a future generator), as presented in Fig.~\ref{fig:intro}(c). However, this approach fails to leverage the potential interactions between understanding and generation. Moreover, the separate processing of these tasks hinders the optimization process, resulting in suboptimal performance. To address this, we propose to initialize a set of world queries using the raw BEV features (before LLM processing). These queries are then enhanced with world knowledge from text tokens through causal attention in the LLM. As a result, by using the world-knowledge-enhanced queries to interact with the LLM-processed BEV features via a \textit{current to future link}, we ensure that the generated scene evolutions are enriched with world knowledge, effectively bridging the gap between generation and understanding.

By consolidating 3D scene understanding and future scene generation within a single framework, \ours~establishes a unified representation that seamlessly accommodates both tasks, offering a holistic perspective on driving environments. This marks a significant step toward a unified DWM, demonstrating the feasibility of integrated driving understanding and generation. Extensive experiments validate the effectiveness of our \ours~in terms of both tasks. Notably, our method significantly reduces the error by 32.4\% compared to the current state-of-the-art (SOTA) method~\cite{yang2024visual} for generation. Additionally, for the understanding task, our approach outperforms the SOTA~\cite{wang2024omnidrive} by 8.0\% under the CIDEr metric on the challenging OmniDrive-nuScenes dataset~\cite{wang2024omnidrive}.

Our major contributions can be summarized as follows: \textbf{1)} In this paper, we propose \ourmethod, which tames the LLM to understand the autonomous driving scene and predict its evolutions simultaneously. To the best of our knowledge, this is the first world model that can unify the 3D understanding and generation task; \textbf{2)} We introduce world queries to capture and integrate world knowledge from text tokens, ensuring that the generated scene evolutions are not only contextually aware but also enriched with world knowledge. This scheme effectively bridges the gap between the understanding and generation tasks, enabling a more coherent and accurate prediction of future scenes.

\section{Related Work}
\label{sec:related Work}
\paragraph{World Models for Driving.} Driving World Models (DWMs)~\cite{ha2018world} have gained considerable attention in autonomous driving for obtaining comprehensive environmental representation and predicting future states based on action sequences. Current research mainly focuses on the generation, whether in 2D~\cite{wang2023drivedreamer,ma2024unleashing,zheng2024doe} or 3D~\cite{min2024driveworld,ma2024cam4docc}. 

Specifically, most pioneering 2D world models perform a video generation for driving scenarios. GAIA-1~\cite{hu2023gaia} first introduced a learned simulator based on an autoregressive model. Recent work further leverages the large scale of data~\cite{yang2024generalized,jia2023adriver,zhang2024bevworld} and more powerful pre-training models, significantly enhancing generation quality regarding consistency~\cite{wang2024driving,gao2023magicdrive}, resolution~\cite{gao2024vista,jia2023adriver}, and controllability~\cite{lu2025wovogen,zhao2024drivedreamer,wen2024panacea,li2025drivingdiffusion}. Concurrently, some studies aim to generate 3D spatial information for future scenes to provide geometric representations that can benefit autonomous driving systems. Occworld~\cite{zheng2025occworld} focuses on future occupancy generation and ego planning using spatial-temporal transformers, which has been adapted to other paradigms including diffusion~\cite{wang2024occsora,gu2024dome}, rendering~\cite{agro2024uno,huang2025neural,yan2024renderworld}, and autoregressive transformer~\cite{wei2024occllama}. Additionally, some approaches~\cite{zyrianov2024lidardm,weng2022s2net,khurana2023point,zhang2023learning} propose future point cloud forecasting as a world model, among which, ViDAR~\cite{yang2024visual} using images to predict future point clouds through a self-supervised manner.

However, existing DWMs overlook the explicit understanding capacity of the driving environment. This paper aims to propose a unified world model that can both comprehend the scenario and generate scene evolution.

\paragraph{Large Language Models for Driving.}
Large Language Models (LLMs) exhibit impressive generalization and extensive world knowledge derived from vast data, showcasing remarkable capabilities across various tasks~\cite{wu2024visionllm,zhang2025psalm,dong2023dreamllm}. This has led researchers to explore their applications in autonomous driving, and current studies~\cite{wang2023drivemlm,shao2024lmdrive,nie2025reason2drive,ding2024holistic} primarily focus on using LLMs to understand driving scenarios and make perceptual or decision-making outputs. For instance, DriveGPT4~\cite{xu2024drivegpt4} processes front-view video input to predict vehicle actions and provide justifications via an LLM. DriveLM~\cite{sima2025drivelm} leverages LLMs for graph-based visual question-answering (VQA) and end-to-end driving. ELM~\cite{zhou2025embodied} enhances the spatial perception and temporal modeling through space-aware pre-training and time-aware token selection. OmniDrive~\cite{wang2024omnidrive} introduces a benchmark with extensive VQA data labeled by GPT-4~\cite{achiam2023gpt} and utilizes Q-Former to integrate 2D pre-trained knowledge with 3D spatial. Despite significant advancements and the emergence of various language-based methods, the application of LLMs in driving remains mainly limited to understanding and text modeling. In this paper, we aim to tame the LLM to understand the autonomous driving scene and predict its future evolutions simultaneously.

\section{Preliminaries}
This section revisits the driving world models and the Bird's Eye View representation as preliminary.

\textbf{The Driving World Models (DWMs)} seek to learn a general representation of the world from large-scale unlabeled driving data by forecasting future scenarios, enabling the model to grasp the data distribution of real situations. Specifically, given an observation $\mathcal{O}_{t}$ at time $t$, the model forecasts information about the next observation $\mathcal{O}_{t+1}$. The framework of DWMs can be summarized as follows:
\begin{eqnarray}
\mathcal{L}_t = \boldsymbol{\mathcal{E}} \left ( \mathcal{O}_{t}\right),
\mathcal{L}_{t+1} = \boldsymbol{\mathcal{M}}\left ( \mathcal{L}_{t}\right), \mathcal{O}_{t+1} = \boldsymbol{\mathcal{D}} \left ( \mathcal{L}_{t+1}\right),
\end{eqnarray}
where $\boldsymbol{\mathcal{E}}$ and $\boldsymbol{\mathcal{D}}$ represent the encoder and decoder for the scene, while the world predictor $\boldsymbol{\mathcal{M}}$ maps the latent state $\mathcal{L}_{t}$ to the next time step $\mathcal{L}_{t+1}$. Together, these components follow the workflow of $\mathcal{O}_{t} \to \mathcal{L}_t \to \mathcal{L}_{t+1} \to \mathcal{O}_{t+1}$.

\textbf{Bird’s-Eye View (BEV)} has emerged recently as a unified representation offering a natural candidate view. The BEVFormer series~\cite{yang2023bevformer,li2022bevformer} exemplifies this approach by leveraging cross-attention to improve 3D-2D view transformation modeling, resulting in robust BEV representations. This BEV feature maintains geometric spatial relationships between views, enabling the model to capture interactions among objects and agents from various perspectives. Additionally, BEV representations are ideal for integrating visual semantics and surrounding geometry, making them well-suited for understanding generation unification.

This paper focuses on unifying the 3D scene understanding and generation by using current multi-view as the observation to generate future point clouds, which inherently maintain accurate geometric relationships among objects and their environments within a BEV-based representation.

\section{\ours}
\begin{figure*}[t]
	\begin{center}
	\includegraphics[width=0.98\linewidth]{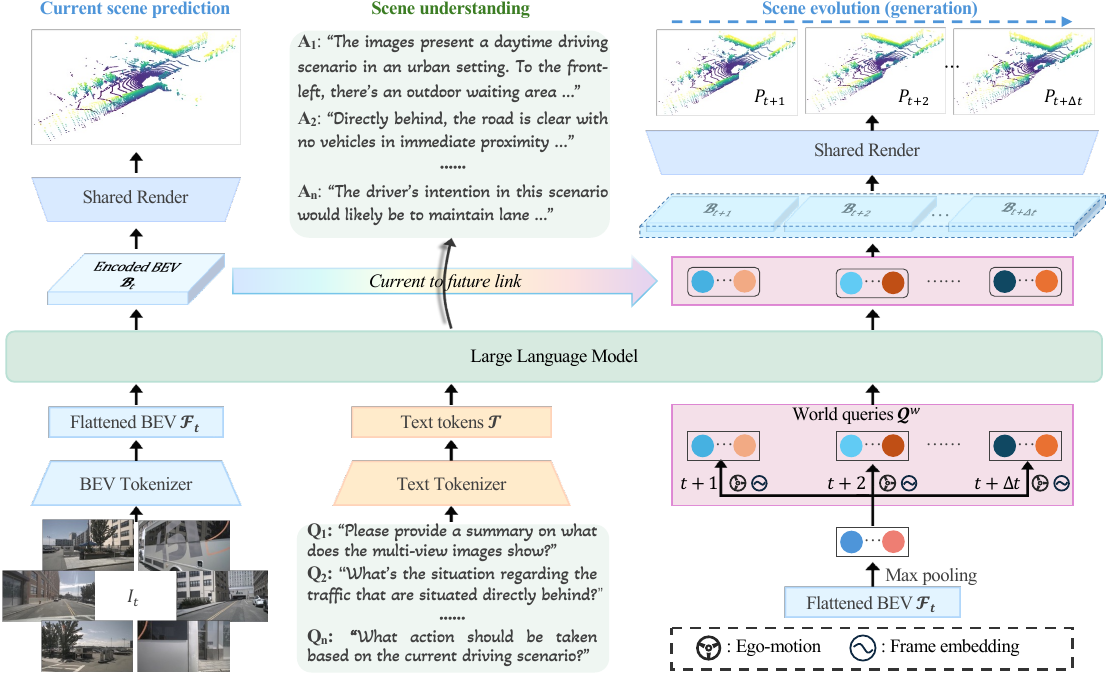}
	\end{center}
        \vspace{-15pt}
	\caption{
    The pipeline of our \ours. The BEV tokenizer converts multi-view $I_{t}$ into flattened BEV $\boldsymbol{\mathcal{F}}_{t}$, which are fed into the large language model (LLM). The LLM interprets user instructions $\boldsymbol{\mathcal{T}}$ and generates textual responses by leveraging its understanding of driving scenes as world knowledge. A group of world queries $\boldsymbol{\mathcal{Q}}^{w}$ are appended to the LLM input sequence. Encoded BEV $\boldsymbol{\mathcal{B}}_{t}$ and world queries generate future BEV ($\boldsymbol{\mathcal{B}}_{t+1},\cdots,\boldsymbol{\mathcal{B}}_{t+\Delta t}$) via a \textit{current to future link}, and the shared Render generates point clouds evolution.
}
\label{fig:pipeline}
\vspace{-10pt}
\end{figure*}
This paper presents \ours, a unified framework for driving scenarios understanding and generating. Our \ours~serving as a world model that predicts scenes' point cloud evolution based on image observations and facilitates detailed scene comprehension. The pipeline of our method is illustrated in Fig.~\ref{fig:pipeline}. We begin with multi-view input images $I_{t}$, which are encoded for semantic information using a BEV-based tokenizer and then processed by a Large Language Model (LLM). The LLM predicts the next token based on user instructions to interpret the current autonomous driving scenario. We integrate world queries into the sequence to transfer world knowledge from the conversations to the generation task. The \textit{current to future link} generates future BEV features, and the shared render predicts the current scene point cloud $P_{t}$ (as an auxiliary task) and generates $\Delta t$ future scenes from $P_{t+1}$ to $P_{t+\Delta t}$.

\subsection{World Tokenizer and Render}
\label{sec:tokenizer}
The world tokenizer encodes the world observation, i.e., the current multi-view images, into a compressed continuous BEV representation, which is then further processed by the LLM. Conversely, the Render~\cite{yang2024unipad,zhu2023ponderv2,wang2021neus} converts BEV features into point clouds to generate geometric information about the scenario. Both modules are detailed as follows.

\noindent\textbf{BEV-based World Tokenizer.} To preserve geometric spatial relationships between views and rich semantic information for LLM inputs, we adopt a BEV-based world tokenizer $\mathcal{E}$. Specifically, the multi-view images $I_{t}$ at time $t$ are passed through a CLIP image encoder~\cite{cherti2023reproducible,liu2022convnet} and a single frame BEVFormer v2~\cite{yang2023bevformer} without modification. The obtained BEV feature $\boldsymbol{\mathcal{F}}^{bev}_{t}\in \mathbb{R}^{w\times h\times c}$ captures the semantic and geometric information, where $w$ and $h$ denote the scale of encoded scene, with larger values indicating greater detail, and $c$ is the channel dimension of BEV. However, such a feature, often containing tens of thousands of tokens, is too large for the LLM~\cite{zhang2024make}. To address this, we implement a down-sampling block that reduces $\boldsymbol{\mathcal{F}}^{bev}_{t}$ by two times, resulting in a compressed shape of $\mathbb{R}^{\frac{w}{4}\times \frac{h}{4}\times (c\times4)}$. When the LLM is required, the down-sampled feature will be flattened and projected to $\boldsymbol{\mathcal{F}}_{t}\in \mathbb{R}^{L_{bev}\times C}$, where $L_{bev} = \frac{w}{4}\times \frac{h}{4}$.

\noindent\textbf{BEV-to-Point Render.}
We introduce a simple BEV-to-Point Render $\mathcal{R}$, aiming to map the aforementioned down-sampled feature to the scene point cloud $P_{t}$. Specifically, we first up-sample the compressed BEV feature (or encoded BEV $\boldsymbol{\mathcal{B}}_{t}\in \mathbb{R}^{L_{bev}\times (c\times4)}$ after processing of the LLM and an out-projection) to the shape of $\mathbb{R}^{w\times h \times c}$ using nearest neighbor interpolation and convolutions. To address the absence of height information in the BEV feature, we reshape the input to $\mathbb{R}^{w \times h \times z \times \frac{c}{z}}$ by adding an extra height dimension. We then apply a series of 3D convolutions to reconstruct the volumetric feature $\boldsymbol{\mathcal{F}}^{vol}_{t}\in \mathbb{R}^{w\times h\times z\times c'}$, where $z$ is the height and $c'$ is the output channel dimension. Finally, we construct rays $\{\mathbf{r}_{k}\}_{k=1}^K$ according to the LiDAR setup of the dataset and use differentiable volume rendering to compute the depth for each ray.

The rendering process models the environment as an implicit signed distance function (SDF) field to capture intricate geometric details accurately~\cite{yang2024unipad,zhu2023ponderv2,wang2021neus}. Given a ray $\mathbf{r}_k$ originating from $\mathbf{o}$ and directed along $\mathbf{t}_k$, we discretize it into $n$ sampled points $\{\mathbf{p}_{i} = \mathbf{o} + d_{i} \mathbf{t}_k \mid i=1,\cdots,n \text{ and } 0\le d_{i}<d_{i+1}\}$, where $\mathbf{p}_{i}$ corresponds to a location in 3D space, determined by its depth $d_i$ along the ray. For each sampled point, we retrieve a local feature embedding $\mathbf{f}_i$ from the volumetric representation $\boldsymbol{\mathcal{F}}^{vol}_{t}$ via trilinear interpolation. Subsequently, a shallow MLP $\phi_{\mathrm{SDF}}$ is used to predict the SDF value $s_i=\phi_{\mathrm{SDF}}(\mathbf{p}_i,\mathbf{f}_i)$. With the predicted SDF values, the rendered depth $\tilde{d}(\mathbf{r}_k)$ is computed through a weighted integration of all sampled depths by $\tilde{d}(\mathbf{r}_k)=\sum_{i=1}^n w_i d_i$, where $w_i=T_i\alpha_i$~\cite{wang2021neus} represents an unbiased, occlusion-aware weight. The transmittance $T_i=\prod_{j=1}^{i-1}(1-\alpha_j)$ accumulates the survival probability of photons up to the $j$-th sample, $\alpha_i=\max(\frac{\sigma_t(s_i)-\sigma_t(s_{i+1})}{\sigma_t(s_i)},0 )$ indicates the opacity, and $\sigma_t(x)=(1+e^{-tx})^{-1}$ is a sigmoid modulated by a learnable parameter $t$.

\subsection{Unification}
\label{sec:unification}
This section introduces the unification of world understanding and future scene generation within our \ours. The Large Language Model (LLM) interprets driving scenarios from world tokenizer outputs ($\boldsymbol{\mathcal{F}}_{t}$) based on user instructions. $\Delta t$ groups of world queries gather knowledge from conversations, aiding in generating scene evolution.

\textbf{Large Language Model.}
\label{sec:llm}
As shown in Fig.~\ref{fig:pipeline}, the LLM is pivotal to our \ours, modeling BEV inputs $\boldsymbol{\mathcal{F}}_{t}$, parsing user instructions, acquiring world knowledge from real driving scenario inquiries, and generating predictions. We utilize the LLM within the widely used InternVL2~\cite{chen2024far}.

\textbf{Understanding.}
Following prior work~\cite{liu2024visual,liu2024improved}, we project the flattened BEV to a shape of $\mathbb{R}^{L_{bev}\times C}$ and into the feature space of the LLM using a two-layer MLP, where $L_{bev}$ is the input BEV length, and $C$ is the channel dimension of LLM. For prompts on the current scene, we tokenize them into distinct vocabulary indices and text tokens $\boldsymbol{\mathcal{T}}$ for processing by the LLM. Like existing multi-modal language models~\cite{li2024monkey,chen2024far}, \ours~responds to user queries about the driving environment, providing scene descriptions and answers to visual questions. The LLM understands the scene through a next-token prediction approach.

\textbf{Generation.} Predicting and generating future changes based on observations of the current moment requires the model to have an exhaustive understanding of the world. To endow the LLM with future-generation capability, we propose a world query technique, which links world knowledge to future scenarios and improves information transfer between the LLM and the Render. As in Fig.~\ref{fig:pipeline}, we outline the generation process in terms of LLM input and output.

For the input to the LLM, we utilize $\Delta t$ groups of world queries $\boldsymbol{\mathcal{Q}}^{w}\in \mathbb{R}^{(\Delta t\times n)\times C}$, where $n$ is the number of queries per group and $C$ represents the channel dimension of LLM. We emphasize the importance of proper feature initialization for effective learning. Thus, we employ a max pooling to derive the world queries from the peak of the BEV feature $\boldsymbol{\mathcal{F}}_{t}$, yielding $\boldsymbol{\mathcal{Q}}\in \mathbb{R}^{n\times (c\times4)}$. The $\boldsymbol{\mathcal{Q}}$ is then copied $\Delta t$ times as query groups $\left \{ \boldsymbol{\mathcal{Q}}_{i} |i=1,\cdots,\Delta t\right \} $. To further enable controllable future generation, we encode the ego-motion condition to $e_{t+i}$, which describes the planned future positions and heading of the ego-vehicle from the current to the $i$-th frame into high-dimensional embeddings. The ego-motion information $e_{t+i}$ is then added to the corresponding queries $\boldsymbol{\mathcal{Q}}_{i}$. Additionally, a frame embedding $\mathrm{FE} \in \mathbb{R}^{\Delta t\times (c\times4)}$ is incorporated by the broadcast mechanism to denote the prediction frames for which group of world queries is responsible. The world queries $\boldsymbol{\mathcal{Q}}^{w}$ and flattened BEV $\boldsymbol{\mathcal{F}}^{t}$ share the language-space projection layer (i.e., MLP) to project from $c\times 4$ to $C$ for the language model channel. The $\boldsymbol{\mathcal{Q}}^{w}\in \mathbb{R}^{(\Delta t\times n)\times C}$ can be computed as:
\begin{equation}
\begin{array}{c}
\hspace{-3.75mm}
\resizebox{.9\hsize}{!}{$\boldsymbol{\mathcal{Q}}^{w} = \mathrm{MLP}(\mathrm{Concat}\left [ \boldsymbol{\mathcal{Q}}_{i}+e_{t+i} \mid  i= \left \{ 1,\cdots,\Delta t \right \} \right] + \mathrm{FE}).$}
\end{array}
\end{equation}

The LLM’s causal attention mechanism (i.e., the later token can access the earlier information) allows world queries to access world knowledge derived from the understanding process. After being processed by the LLM feed-forwardly, the encoded BEV feature and world quires are projected by a shared two-layer MLP from the channel dimension of LLM $C$ back to the channel of $c \times 4$. Note that each group of world queries contains only $n$ queries, which provide a sparse view of the future world, complicating the reconstruction of the future scene by the world Render. To address this, we propose the \textit{current to future link} module, which employs cross-attention layers to inject world knowledge for future BEV features. Specifically, the \textit{current to future link} module contains $3$ cross-attention blocks for generating future BEV features. Each cross-attention block includes a cross-attention layer that uses the encoded BEV $\boldsymbol{\mathcal{B}}_{t}$ from the LLM output as the query, with world queries for each scene serving as the value and key. A self-attention layer and a feed-forward network further process spatial information. The encoded BEV ($\boldsymbol{\mathcal{B}}_{t}$) and generated future BEV features ($\boldsymbol{\mathcal{B}}_{t+1},\cdots,\boldsymbol{\mathcal{B}}_{t+\Delta t}$) are sent to a shared world Render and obtain point cloud from $P_{t}$ to $P_{t+\Delta t}$.

\subsection{Training Objectives}
To perform auto-regressive language modeling, we employ Next Token Prediction (NTP) to maximize the likelihood of text tokens, following the standard language objective:
\begin{equation}
\mathcal{L}_{N} = -\sum_{i=1}\log P\left ( \boldsymbol{\mathcal{T}}_{i}|\boldsymbol{\mathcal{F}}_{t},\boldsymbol{\mathcal{T}}_{1},\cdots, \boldsymbol{\mathcal{T}}_{i-1}; \boldsymbol{\Theta} \right ),
\end{equation}
where $P\left ( \cdot | \cdot \right ) $ represents the conditional probability modeled by the weights $\boldsymbol{\Theta}$, $\boldsymbol{\mathcal{F}}_{t}$ is the flattened BEV feature for the input frame, and $\boldsymbol{\mathcal{T}}_{i}$ denotes the $i$-th text token. 

For point cloud generation, we supervise the depths of various rays $d(\mathbf{r}_k)$ using only L1 loss:
\begin{equation}
\label{eq:depthloss}
\mathcal{L}_{D} = \sum_{i=0}^{\Delta t}\lambda_{i} \frac{1}{N_{i}} \sum_{k=0}^{N_{i}}\left |d(\mathbf{r}_k) -\tilde{d}(\mathbf{r}_k) \right |,
\end{equation}
where $\lambda_{i}$ is the loss weight for frame $t+i$, and $N_{i}$ is the number of rays in the point clouds for frame $t+i$. The total loss for \ours~is given by $\mathcal{L}=\mathcal{L}_{N}+10\mathcal{L}_{D}$.

\begin{table*}[t]
\setlength{\tabcolsep}{2.5mm}
\centering
\footnotesize
\caption{The comparison of our \ours~and understanding/generation specialist models. L/C/T refers to LiDAR/camera/text, respectively. We report MTETOR, CIDEr, and ROUGE for understanding tasks, and Chamfer distance for 0-3s on the (OmniDrive-)nuScenes validation set, following ViDAR~\cite{yang2024visual}. \textcolor{red}{$^\dag$} denotes results from ViDAR, while scores for GPT-4o and LLaVA-OV are sourced from DriveMM~\cite{huang2024drivemm}.}
\label{tab:main}
\vspace{-10pt}
\begin{tabular}{ lcccccccccc }
   \toprule
 \multirow{2.3}{*}{Method} & \multirow{2.3}{*}{Reference} & \multirow{2.3}{*}{\# LLM Params}& \multirow{2.3}{*}{Modality}&\multicolumn{4}{c}{Generation} & \multicolumn{3}{c}{Understanding}\\
\cmidrule(lr){5-8}\cmidrule(lr){9-11}
 & & & & 0s $\downarrow$ & 1s $\downarrow$ & 2s $\downarrow$ & 3s $\downarrow$ & MTETOR $\uparrow$ & ROUGE $\uparrow$ & CIDEr $\uparrow$ \\
\midrule
\multicolumn{11}{c}{\textit{Only Generation}} \\
\midrule
4D-Occ\textcolor{red}{$^\dag$}~\cite{khurana2023point}& CVPR 23 & -&L$\to$L &-&1.13 &1.53 &2.11&\multicolumn{3}{c}{\multirow{2}{*}{Unsupported}}\\
ViDAR~\cite{yang2024visual}& CVPR 24 & - &C$\to$L&- & 1.12 & 1.38& 1.73&&&\\
\midrule
\multicolumn{11}{c}{\textit{Only Understanding}} \\
\midrule
GPT-4o~\cite{hurst2024gpt}& -& -&C$\to$T&\multicolumn{4}{c}{\multirow{5}{*}{Unsupported}}& -& 0.223 & 0.244\\
LLaVA-OV~\cite{li2024llavaov}& arXiv 24& 7B&C$\to$T&&&&& -& 0.221 & 0.284\\
OmniDrive~\cite{wang2024omnidrive} & CVPR 25& 7B&C$\to$T&&&&& 0.380 & 0.326 & 0.686\\
OmniDrive-2D~\cite{wang2024omnidrive} & CVPR 25& 7B&C$\to$T&&&&& 0.383 & 0.325 & 0.671\\
OmniDrive-BEV~\cite{wang2024omnidrive} & CVPR 25& 7B&C$\to$T&&&&& 0.356 & 0.278 & 0.595\\
\midrule
\multicolumn{11}{c}{\textit{Unified Understanding and Generation}} \\
\midrule
Separated unification & - & 1.8B &C$\to$T\&L& 0.60 & 0.84 & 1.08 & 1.37 & 0.384 & 0.327 & \textbf{0.745}\\
\rowcolor{linecolor}\ours~(\textbf{ours}) & - & 1.8B &C$\to$T\&L& \textbf{0.59} & \textbf{0.78} & \textbf{0.95} & \textbf{1.17} & \textbf{0.384} & \textbf{0.327} & 0.741\\
\bottomrule
\vspace{-15pt}
\end{tabular}
\end{table*}

\section{Experiments}
\subsection{Dataset and Evaluation Metric}
\noindent\textbf{Datasets.}
1) \textbf{NuScenes}~\cite{caesar2020nuscenes} is a widely used autonomous driving dataset, which includes 700 training scenes, 150 validation scenes, and 150 test scenes. We use six images and the point clouds captured by surrounding cameras and the LiDAR, respectively. 2) \textbf{NuInteract}~\cite{zhao2025extending} is a newly proposed language-based driving dataset with dense captions for each image and scene. With $\sim$1.5M annotations, it supports various tasks, such as 2D perception and 3D visual grounding. 3) \textbf{OmniDrive-nuScenes}~\cite{wang2024omnidrive} supplements nuScenes with high-quality caption and visual question-answering (QA) text pairs generated by GPT4. Considering the high quality and rich VQA annotations, we perform the understanding training and evaluation on the OmniDrive-nuScenes description and conversation data.

\noindent\textbf{Evaluation Metric.}
For understanding tasks, we utilize widely used METEOR~\cite{banerjee2005meteor}, CIDEr~\cite{vedantam2015cider}, and ROUGE~\cite{lin2004rouge} metrics to compute similarities between generated and ground-truth answers at the word level. For generation evaluation, we follow previous work~\cite{yang2024visual} and use Chamfer Distance to measure precision in generated point clouds, considering only points within the range of [-51.2m, 51.2m] on the X- and Y-axes, and [-3m, 5m] on the Z-axis.

\subsection{Main Results}

We compare \ours~with understanding~\cite{hurst2024gpt,li2024llavaov,wang2024omnidrive} and generation~\cite{khurana2023point,yang2024visual} specialist models in Tab.~\ref{tab:main}, which demonstrates competitive performance on both tasks and promote strong unification.

\textbf{For future point cloud generation}, both 4D-Occ and ViDAR utilize a 3s history horizon, while our \ours~only relies on the current frame, achieving significant improvements. Remarkably, with only multi-view inputs, \ours~is capable of generating a more accurate representation of the scene geometry for predicting future evolution, resulting in $\sim$32\% Chamfer Distance reduction in 3s point clouds compared to ViDAR. It should be noted that ViDAR utilizes a carefully designed latent render and an FCOS3D~\cite{wang2021fcos3d} pre-trained backbone, while \ours~uses simple volumetric representation. Furthermore, \ours~can simultaneously understand the current scenario, which is a crucial capability for driving systems but is challenging for existing driving world models.

\textbf{For 3D scene understanding}, we continuously achieve highly competitive results in caption quality compared to understanding specialists. For example, we notably outperform OmniDrive by 8\% on the CIDEr metric and excel in MTETOR and ROUGE. Note that OmniDrive leverages extensive 2D pre-training data~\cite{fang2024eva}, supervision from 3D objects~\cite{wang2023exploring}, and lane detection. The OmniDrive-BEV uses LSS~\cite{philion2020lift} to transform perspective features into a BEV feature and SOLOFusion~\cite{park2022time} for temporal modeling. While it utilizes a BEV representation, its scene understanding is limited, likely due to insufficient data for BEV-based image-text alignment. Further compared to the Separated unification model, our unified approach yields significantly better generation results while maintaining strong understanding performance. This demonstrates successful cross-task knowledge transfer and consistent scene modeling within our compact, unified \ours.

\subsection{Ablation Study}
Unless otherwise specified, we perform ablation studies trained on a quarter of the nuScenes training scenes. Default settings are marked in \colorbox{linecolor}{green}.

\noindent\textbf{Analysis on understanding and generation interaction.} Our \ours~achieves a strong and seamless unification of understanding and generation in driving scenes. We first analyze the relationship between these two processes. As shown in Tab.~\ref{tab:ablation-ralation}, we conduct experiments with four approaches: solely understanding, solely generation, a separated unification, and our method. The separated unification involves sharing the flattened BEV while using separate models for understanding and generation, as in Fig.~\ref{fig:intro}(c). We find that \ours~achieves highly competitive results compared to training on one task, with a minor performance gap (e.g., 0.002 difference on MTETOR/ROUGE and a 0.03 chamfer distance gap on the 3s generation). Nevertheless, our approach shows better results at 0-1s, indicating ongoing optimization challenges in the unified understanding and generation stage. Ultimately, our \ours~outperforms the separated unification in generation results, as the latter fails to exploit the potential interactions between understanding and generation, and the separation hinders optimization and leads to suboptimal performance.

\begin{table}[t]
    \centering
    \footnotesize
    \caption{Ablation on interaction of tasks.}
    \vspace{-10pt}
    \label{tab:ablation-ralation}
    \setlength\tabcolsep{0.45mm}
    \begin{tabular}{ccccccccccc}
        \toprule
        \multirow{2.3}{*}{Under.} & \multirow{2.3}{*}{Gen.} & \multicolumn{4}{c}{Generation} & \multicolumn{3}{c}{Understanding}\\
        \cmidrule(lr){3-6}\cmidrule(lr){7-9}
        & & 0s $\downarrow$ & 1s $\downarrow$ & 2s $\downarrow$ & 3s $\downarrow$ & MTETOR $\uparrow$ & ROUGE $\uparrow$ & CIDEr $\uparrow$ \\
        \midrule
        \textcolor{gray}{\checkmark} & \textcolor{gray}{-} & \textcolor{gray}{-} & \textcolor{gray}{-} & \textcolor{gray}{-} & \textcolor{gray}{-} & \textcolor{gray}{0.379} & \textcolor{gray}{0.323} & \textcolor{gray}{0.728} \\
        \textcolor{gray}{-}& \textcolor{gray}{\checkmark}& \textcolor{gray}{0.651} & \textcolor{gray}{0.988} & \textcolor{gray}{1.313} & \textcolor{gray}{1.687}&\textcolor{gray}{-}&\textcolor{gray}{-}&\textcolor{gray}{-} \\
        \midrule
        \multicolumn{2}{c}{\scriptsize Separated unify} & 0.663 & 1.095 & 1.476 & 1.875 & 0.377 & 0.321 & \textbf{0.722} \\
        \rowcolor{linecolor}\checkmark & \checkmark & \textbf{0.645} & \textbf{0.984} & \textbf{1.333} & \textbf{1.718} & \textbf{0.377} & \textbf{0.321} & 0.720 \\
        \bottomrule
    \end{tabular}
    \vspace{-5pt}
\end{table}

\begin{figure}[t]
	\begin{center}
	\includegraphics[width=0.98\linewidth]{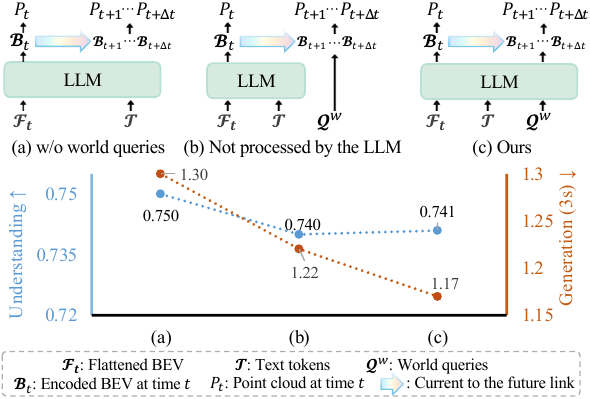}
	\end{center}
        \vspace{-15pt}
	\caption{The effect of world queries for understanding (CIDEr) and generation (3s chamfer distance) is trained on full data.}
	\label{fig:worldquery}
        \vspace{-5pt}
\end{figure}

\noindent\textbf{Analysis on the effect of the world queries.} We then validate the efficacy of world queries ($\boldsymbol{\mathcal{Q}}^{w}$), as shown in Fig.~\ref{fig:worldquery}. Without world queries (Fig.~\ref{fig:worldquery}(a)), the sequences $\boldsymbol{\mathcal{B}}_{t+1}, \cdots, \boldsymbol{\mathcal{B}}_{t+\Delta t}$ are generated by directly integrating future ego motion into $\boldsymbol{\mathcal{B}}_{t}$. It can be found that introducing world queries significantly improves future generation capabilities, reducing the Chamfer Distance for 3s point cloud prediction by 10\%. Although there is a slight 1\% decrease in understanding performance (CIDEr), we attribute this to the increased optimization complexity from adding new informational parameters to the LLM. A comparison of Fig.~\ref{fig:worldquery}(b)-(c) further reveals that $\boldsymbol{\mathcal{Q}}^{w}$ processed through the LLM enhances generative performance, supporting their role in effective world knowledge transfer.

\noindent\textbf{Analysis on generation length.} In Tab.~\ref{tab:ablation-length}, we discuss the number of generated frames for future scenes. As the number increases, the generation results exhibit a slight drop, which we attribute to an optimization dilemma within the LLM. This suggests that a more efficient interaction method could be explored in the future. Additionally, we evaluate the auxiliary role of current point cloud prediction for future generation, as shown in the last two rows of Tab.~\ref{tab:ablation-length}. Predicting the current frame regularizes the encoded BEV ($\boldsymbol{\mathcal{B}_{t}}$) from the LLM outputs, enhancing future generation results. More importantly, training for current point cloud prediction does not add extra inference burden for future generations, serving as a practical auxiliary task.

\begin{table}[t]
    \centering
    \footnotesize
    \caption{Ablation on generation length.}
    \label{tab:ablation-length}
    \setlength\tabcolsep{0.65mm}
    \begin{tabular}{ccccccccccc}
        \toprule
        \multicolumn{4}{c}{Second} &\multicolumn{4}{c}{Generation} & \multicolumn{3}{c}{Understanding}\\
        \cmidrule(lr){1-4}\cmidrule(lr){5-8}\cmidrule(lr){9-11}
        0&1&2 & 3 & 0s $\downarrow$ & 1s $\downarrow$ & 2s $\downarrow$ & 3s $\downarrow$ & MTETOR $\uparrow$ & ROUGE $\uparrow$ & CIDEr $\uparrow$ \\
        \midrule
        \checkmark &\checkmark & -& - & \textbf{0.607} & \textbf{0.944} & - & - & \textbf{0.379} & \textbf{0.323} & \textbf{0.725} \\
        \checkmark &\checkmark & \checkmark& - & 0.632 & 0.951 & \textbf{1.313} & - & 0.378 & 0.321 & 0.714 \\
        - &\checkmark & \checkmark &\checkmark & - & 1.078 & 1.397 & 1.779 & 0.378 & 0.321 & 0.717 \\
        \rowcolor{linecolor}\checkmark &\checkmark & \checkmark &\checkmark & 0.645 & 0.984 & 1.333 & \textbf{1.718} & 0.377 & 0.321 & 0.720 \\
        \bottomrule
    \end{tabular}
\end{table}

\begin{table}[t]
    \centering
    \footnotesize
    \caption{Ablation on the source of world queries.}
    \label{tab:ablation-init}
    \setlength\tabcolsep{0.9mm}
    \begin{tabular}{cccccccc}
        \toprule
        \multirow{2.3}{*}{Pooling} & \multicolumn{4}{c}{Generation} & \multicolumn{3}{c}{Understanding}\\
        \cmidrule(lr){2-5}\cmidrule(lr){6-8}
        & 0s $\downarrow$ & 1s $\downarrow$ & 2s $\downarrow$ & 3s $\downarrow$ & MTETOR $\uparrow$ & ROUGE $\uparrow$ & CIDEr $\uparrow$ \\
        \midrule
        Attn. & 0.656 & 1.001 & 1.344 & 1.748 & 0.377 & 0.321 & 0.712 \\
        Avg. & 0.660 & 0.996 & 1.348 & 1.741 & 0.376 & 0.321 & 0.715 \\
        \rowcolor{linecolor}Max & \textbf{0.645} & \textbf{0.984} & \textbf{1.333} & \textbf{1.718} & \textbf{0.377} & \textbf{0.321} & \textbf{0.720} \\
        \bottomrule
    \end{tabular}
\end{table}

\begin{table}[t]
    \centering
    \footnotesize
    \caption{Ablation on size for the flattened BEV.}
    \label{tab:ablation-bevsize}
    \setlength\tabcolsep{0.8mm}
    \begin{tabular}{cccccccccc}
        \toprule
        \multirow{2.3}{*}{BEV size} & \multicolumn{4}{c}{Generation} & \multicolumn{3}{c}{Understanding}\\
        \cmidrule(lr){2-5}\cmidrule(lr){6-8}
        & 0s $\downarrow$ & 1s $\downarrow$ & 2s $\downarrow$ & 3s $\downarrow$ & MTETOR $\uparrow$ & ROUGE $\uparrow$ & CIDEr $\uparrow$ \\
        \midrule
        25 $\times$ 25 & 0.720 & 1.040 & 1.347 & \textbf{1.698} & 0.367 & 0.311 & 0.671\\
        \rowcolor{linecolor} 50 $\times$ 50 & \textbf{0.645} & \textbf{0.984} & \textbf{1.333} & 1.718 & \textbf{0.377} & \textbf{0.321} & \textbf{0.720}\\
        \bottomrule
    \end{tabular}
\end{table}

\noindent\textbf{Analysis on the source of world queries.} We also assess the initialization of world queries in Tab.~\ref{tab:ablation-init}, including attention pooling~\cite{vaswani2017attention,radford2021learning}, adaptive average pooling and max pooling. It shows that world queries derived from the adaptive max pool of the flattened BEV $\boldsymbol{\mathcal{F}}_{t}$ perform better in chamfer distance on 3s. We argue that the max pool effectively captures peak responses from $\boldsymbol{\mathcal{F}}_{t}$, whereas average or attention pooling may overly emphasize global information, potentially affected by background noise.

\noindent\textbf{Analysis on the size of flattened BEV.} We finally analyze the impact of flattened BEV ($\boldsymbol{\mathcal{F}}_{t}$) size by varying the BEV tokenizer's downsampling multiplier, resulting in different feature lengths. Experiments with 8$\times$ and 4$\times$ downsampling yield $\boldsymbol{\mathcal{F}}_{t}$ with spatial resolutions of 25 and 50, respectively. Tab.~\ref{tab:ablation-bevsize} demonstrates that a spatial size of $50 \times 50$ significantly improves CIDEr and 0s generation by 7.3\% and 10\% compared to $25 \times 25$. We attribute this improvement to reduced information loss compared to excessive downsampling, enhancing text comprehension and facilitating point cloud recovery/prediction. While further increasing the size of $\boldsymbol{\mathcal{F}}_{t}$ might improve performance, we chose $50 \times 50$ as a trade-off due to LLM processing length limitations.

\begin{figure*}[t]
    \begin{center}
	\includegraphics[width=\linewidth]{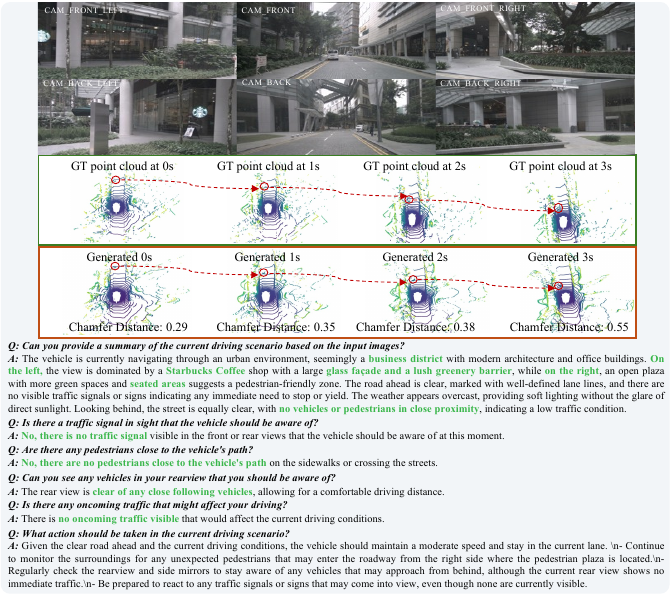}
	\end{center}
        \vspace{-15pt}
    \caption{Qualitative results for future generation and scene understanding. From top to bottom, we display the multi-view input of the current scene, the ground truth scene evolution, the generated scene evolution, and the scene understanding result.}
    \label{fig:demo}
    \vspace{-10pt}
\end{figure*}

\subsection{Qualitative Results}
This section presents qualitative results on future generations and scene understanding, as illustrated in Fig.~\ref{fig:demo}. Our \ours~effectively captures future scene evolution, such as the van keeping close to the ego vehicle, noted in the red circle. Furthermore, \ours~exhibits a strong understanding of BEV inputs, as indicated by the green text, accurately identifying objects such as ``Starbucks".

\section{Conclusion}

This paper introduces \ourmethod, a simple yet effective unified Driving World Model that integrates 3D scene understanding and future scene generation within a single framework. We effectively bridge the gap between understanding and generation by leveraging a Bird’s-Eye View representation and incorporating world queries enhanced through large language models. Extensive experiments validate the effectiveness of the proposed \ours, demonstrating significant improvements in future scene prediction accuracy and understanding metrics. We believe this work is a significant step toward a stronger unified Driving World Model.

\noindent\textbf{Limitation.} Although \ours~achieves promising results in unifying 3D scene understanding and generation, there are some limitations: 1) We have not explored perception tasks for autonomous driving within our framework. 2) Future image is also an important generation modality but is still under exploration. We left these in our future work.

\noindent\textbf{Acknowledgment.} This work was supported by the NSFC (62225603, 62441615, and 623B2038).

\maketitlesupplementary
\setcounter{figure}{0}
\setcounter{table}{0}

\setcounter{section}{0}
\renewcommand{\thesection}{S\arabic{section}}
\renewcommand\thefigure{S\arabic{figure}}
\renewcommand\thetable{S\arabic{table}}

\section{Additional Experiments}

\subsection{Training Details}
\label{sec:training-stage}
The BEV-based tokenizer utilizes the OpenCLIP ConNext-L backbone~\cite{liu2022convnet, cherti2023reproducible, radford2021learning}, while other modules in the tokenizer and Render are trained from scratch. The LLM is derived from InternVL2-2B~\cite{chen2024internvl, chen2024far}. The resolution of the input image is $1600\times 900$, while the BEV-based world tokenizer adopts the same hyperparameters as BEVFormer v2-base~\cite{yang2023bevformer}, with the size of the encoded scene set to $w=h=200$ and a BEV channel dimension of 256. The $z$ and $c'$ in the BEV-to-point clouds Render are set to $32$. For future generation, we forecast scene evolution over 3 seconds, i.e., $\Delta t = 3$. The frame-wise weights in Eq.~\ref{eq:depthloss} of the main paper are empirically defined by $\lambda_i = 1 + 0.5 \times i, i \in \{0, \cdots, 3\}$, corresponding to the point clouds from 0 to 3s. The training of \ours~is structured into three stages and detailed below. Additional details are provided in Tab.~\ref{tab:details}.

\noindent{\textbf{Stage-1: Tokenizer Traning.}} In initial stage, we train the world tokenizer $\mathcal{E}$ and Render $\mathcal{R}$ to convert current images ($I_{t}$) into point clouds ($P_{t}$), following $P_{t} = \mathcal{R} \left ( \mathcal{E} \left ( I_{t}\right ) \right )$. We utilize 12Hz data from the nuScenes training set for the tokenizer and Render learning. 

\noindent{\textbf{Stage-2: BEV-Text Alignment and Refinement.}} This stage encompasses BEV-Text alignment and refinement tuning phases. The alignment phase aims to establish vision-language alignment between the input and output BEV of the LLM, training only the in-projections for flattened BEV embeddings and out-projections for the encoded BEV. To alleviate data deprivation, we propose a simple data augmentation involving masking one of the multi-view images, splicing the caption from the visible view, and using the unprocessed multi-view scene descriptions. This approach increases the multi-view image-text pairs to $\sim$200K, a sevenfold increase from the nuScenes keyframes. In the refinement phase, all parameters are unfrozen, and the LLM is fine-tuned using LoRA~\cite{hu2021lora}. The alignment phase employs NuInteract~\cite{zhao2025extending} dense caption data, while the refinement phase adapts labeling styles using scene description data from OmniDrive-nuScenes~\cite{wang2024omnidrive}.

\noindent{\textbf{Stage-3: Understanding and Generation Unification.}} Building on the understanding gained in the first two stages, we introduce future generation modules to generate point clouds at different moments. We train using nuScenes keyframes, descriptions, and general conversation annotations from OmniDrive-nuScenes.

\begin{table}[!t]
\setlength{\tabcolsep}{3.2mm}
\centering
\caption{Training details of \ours. -/- in Stage 2 indicates BEV-text alignment/refinement.
}
\footnotesize
\label{tab:details}
\begin{tabular}{lccc}
 \toprule
 Config & Stage 1 & Stage 2 & Stage 3\\
 \midrule
 Optimizer & AdamW & AdamW & AdamW\\
 Learning Rate & 2e-4 & 2e-4/4e-4 & 4e-4\\
 Training Epochs & 6 & 3/6 & 36\\
 Learning Rate Scheduler & Cosine & Cosine & Cosine\\
 Batch Size Per GPU & 1 & 4 & 4\\
GPU Device & \multicolumn{3}{c}{32$\times$NVIDIA H20} \\ 
\bottomrule
\end{tabular}
\end{table}

\begin{table}[t]
    \centering
    \footnotesize
    \caption{Ablation on scaling potential of the LLM.}
    \label{tab:ablation-llm}
    \setlength\tabcolsep{0.37mm}
    \begin{tabular}{cccccccccc}
        \toprule
        \multirow{2.3}{*}{\# LLM Params} & \multicolumn{4}{c}{Generation} & \multicolumn{3}{c}{Understanding}\\
        \cmidrule(lr){2-5}\cmidrule(lr){6-8}
        & 0s $\downarrow$ & 1s $\downarrow$ & 2s $\downarrow$ & 3s $\downarrow$ & MTETOR $\uparrow$ & ROUGE $\uparrow$ & CIDEr $\uparrow$ \\
        \midrule
        0.8B & 0.668 & 1.015 & 1.379 & 1.809 & 0.372 & 0.318 & 0.703\\
        \rowcolor{linecolor}1.8B & 0.645 & \textbf{0.984} & 1.333 & 1.718 & 0.377 & 0.321 & 0.720\\
        3.8B & \textbf{0.643} & 0.991 & \textbf{1.321} & \textbf{1.701} & \textbf{0.381} & \textbf{0.325} & \textbf{0.730}\\
        \bottomrule
    \end{tabular}
\end{table}

\begin{table}[!t]
\setlength{\tabcolsep}{1.25mm}
\centering
\caption{Ablation on the number $n$ of world queries.}
\footnotesize
\label{tab:ablation-n}
\begin{tabular}{cccccccc}
    \toprule
    \multirow{2.3}{*}{$n$} & \multicolumn{4}{c}{Generation} & \multicolumn{3}{c}{Understanding}\\
    \cmidrule(lr){2-5}\cmidrule(lr){6-8}
    & 0s $\downarrow$ & 1s $\downarrow$ & 2s $\downarrow$ & 3s $\downarrow$ & MTETOR $\uparrow$ & ROUGE $\uparrow$ & CIDEr $\uparrow$ \\
    \midrule
    1  & 0.658 & 0.996 & 1.328 & 1.725 & 0.376 & 0.320 & 0.712 \\
    2  & 0.656 & 0.995 & \textbf{1.324} & 1.720 & 0.377 & 0.321 & 0.714 \\
    \rowcolor{linecolor}4  & \textbf{0.645} & \textbf{0.984} & 1.333 & \textbf{1.718} & 0.377 & \textbf{0.321} & \textbf{0.720} \\
    8  & 0.667 & 1.028 & 1.361 & 1.744 & 0.376 & 0.321 & 0.713 \\
    16 & 0.658 & 0.999 & 1.354 & 1.748 & \textbf{0.378} & 0.321 & 0.716 \\
    \bottomrule
\end{tabular}
\end{table}

\subsection{Additional Ablation Study}
Unless otherwise specified, we perform ablation studies trained on a quarter of the nuScenes training scenes. Default settings are marked in \colorbox{linecolor}{green}.

\noindent\textbf{Analysis on the scaling potential of the LLM.} We first explore the scaling potential of our \ours, as shown in Tab.~\ref{tab:ablation-llm}. Scaling up LLMs yields consistent gains in 3D scene understanding and point cloud generation, and we utilize the 1.8B LLM form InternVL2-2B~\cite{chen2024internvl, chen2024far} as a trade-off. This indicates that the broader world knowledge acquired during pre-training enhances these tasks, suggesting potential benefits from further scaling.
\begin{figure*}[t]
	\begin{center}
		\includegraphics[width=0.95\linewidth]{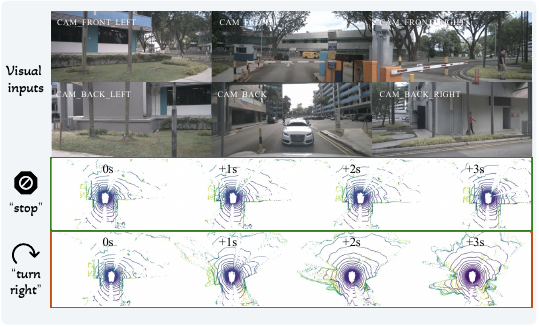}
	\end{center}
    \vspace{-10pt}
	\caption{Qualitative results of \ours~conditioned on different future ego-motion conditions. From top to bottom, each sub-figure displays the multi-view input of the current scene, scene evolution predicted with a ``stop" future ego-motion, and scene evolution predicted with a ``turn right" ego-motion.}
	\label{fig:control}
        \vspace{-10pt}
\end{figure*}

\begin{table}[t]
    \setlength{\tabcolsep}{0.9mm}
    \centering
    \caption{Comparison of generation ability.}
    \vspace{-5pt}
    \footnotesize
    \label{tab:ablation-ability}
    \begin{tabular}{lccccccc}
        \toprule
        \multirow{2.3}{*}{} & \multicolumn{4}{c}{Generation} & \multicolumn{3}{c}{Understanding}\\
        \cmidrule(lr){2-5}\cmidrule(lr){6-8}
        & 0s $\downarrow$ & 1s $\downarrow$ & 2s $\downarrow$ & 3s $\downarrow$ & MTETOR $\uparrow$ & ROUGE $\uparrow$ & CIDEr $\uparrow$ \\
        \midrule
        Copy\&Paste & - & 1.27 & 2.12 & 2.66 & - & - & - \\
        ViDAR~\cite{yang2024visual} & - & 1.12 & 1.38 & 1.73 & - & - & - \\
        \rowcolor{linecolor}\ours & \textbf{0.59} & \textbf{0.78} & \textbf{0.95} & \textbf{1.17} & \textbf{0.384} & \textbf{0.327} & \textbf{0.741}  \\
        \bottomrule
    \end{tabular}
\end{table}

\begin{table}[t]
\footnotesize
\centering
\setlength{\tabcolsep}{2.3mm}
\caption{VQA results on NuScenes-QA.}
\vspace{-5pt}
\label{tab:nusc_qa}
\begin{tabular}{lcccccccccc}
\toprule
Method & Reference & Modality & Acc. (\%) $\uparrow$ \\
\midrule
 LLaVA~\cite{liu2023visual} & NeurIPS 23 & Camera & 47.4 \\
 LiDAR-LLM~\cite{yang2023lidar} & arXiv 23 & LiDAR & 48.6 \\
 BEVDet+BUTD~\cite{qian2024nuscenes} & AAAI 24 & Camera & 57.0 \\
 BEVDet+MCAN~\cite{qian2024nuscenes} & AAAI 24 & Camera & 57.9 \\
 CenterPoint+BUTD~\cite{qian2024nuscenes} & AAAI 24 & LiDAR & 58.1 \\
 CenterPoint+MCAN~\cite{qian2024nuscenes} & AAAI 24 & LiDAR & 59.5 \\
 OmniDrive~\cite{wang2024omnidrive} & CVPR 25 & Camera & 59.2 \\ 
\midrule
\rowcolor{linecolor}\ours & - & Camera & \textbf{61.9}\\
\bottomrule
\end{tabular}
\end{table}

\noindent\textbf{Analysis on the number of world queries.} The world queries facilitate knowledge transfer between the LLM and the Render for future scenarios. We then evaluate the impact of the number of queries $n$ for each group, as shown in Tab.~\ref{tab:ablation-n}. We find that world queries do not adversely affect text understanding quality. However, increasing the number of world queries leads to a decline in performance, likely due to redundant information and optimization challenges. Therefore, we choose to include four world queries per group for future generations.

\noindent\textbf{Analysis on generation ability.} We finally compare our future point cloud generation ability trained on the full training set against a Copy\&Paste baseline, where Copy\&Paste simply duplicates the current ground-truth point cloud for future observations. As shown in Tab.~\ref{tab:ablation-ability}, this baseline fails to account for point cloud changes due to movement and occlusion, demonstrating that \ours~truly learns to understand 3D scenes and predict their future evolution.

\subsection{Understanding on NuScenes-QA}
The NuScenes-QA~\cite{qian2024nuscenes} is another multi-modal VQA benchmark for driving scenarios, featuring primarily single-word answers focused on perception. We fine-tune \ours~on the NuScenes-QA training set to align with its style and length, and the results are shown in Tab.~\ref{tab:nusc_qa}. \ours~achieves superior performance, outperforming LLaVA~\cite{liu2023visual} by 14.5\% and the point cloud method CenterPoint+MCAN~\cite{qian2024nuscenes} by 2.4\%. This showcases \ours's strong 3D scene understanding capabilities via its unified BEV representation, especially considering it requires no 3D object detection supervision.

\section{Discussion}

The integration of Bird's-Eye View (BEV) representations as input for Large Language Models (LLMs) presents distinct advantages in our \ours. Unlike conventional multi-view processing approaches that process individual camera streams independently, the BEV-based tokenization establishes a unified spatial coordinate system that inherently preserves geometric relationships across views while maintaining object interaction patterns. This spatial consolidation addresses the inherent limitations of vision-language models in interpreting multi-perspective scenarios, where disconnected 2D projections fail to capture the holistic 3D environment context. By strategically compressing high-resolution multi-view inputs (1600×900 per view, for example) into a compact BEV latent space through our downsampling block, we achieve efficient token utilization (2,500 tokens vs. $\sim$47,000 tokens for raw view processing) without exceeding standard LLM context windows. Crucially, the spatial-aware BEV features enable synergistic knowledge transfer between scene understanding and generation tasks through our world query mechanism, i.e., the positional correspondence between text descriptions and geometric features permits causal attention patterns that enrich future predictions with linguistic context. Our experiments on nuScenes demonstrate that this spatial-textual alignment contributes substantially to the 32.4\% reduction in generation error and 8.0\% CIDEr improvement, validating BEV's dual role as both information compressor and cross-modal interface.

\newpage
\section{More Qualitative Results}
This section presents further qualitative results of \ours~on controllability and the unification ability of understanding and generation.

\textbf{Potential for Controlled Scene Generation.} As shown in Fig.~\ref{fig:control}, we observe the capability of \ours~to generate future point cloud evolution conditioned on specific ego-motion information, such as ``stop" or ``turn right". This showcases the potential of \ours~as world simulator and its ability to understand complex world scenarios deeply.

\textbf{Unification of understanding and generation.} More qualitative results on future generations and scene understanding are illustrated in Fig.~\ref{fig:vis}. Our \ours~effectively captures future scene evolution (with the ground truth ego-motion information for better comparison), such as the corner of the building keeps moving backward, noted in the red circle in Fig.~\ref{fig:vis3}. While \ours~achieve an encouraging integration of understanding and generation, it faces challenges in complex scenes (e.g., significant left turns and occlusions as in Fig.~\ref{fig:vis2}) and low-quality visible light conditions (e.g., nighttime driving as in Fig.~\ref{fig:vis4}). Despite the complexity of the scenarios, \ours~still makes reasonable predictions about the emerging parts of future scenes.
\begin{figure*}[ht]
    \begin{subfigure}{\linewidth}
        \centering
        \includegraphics[width=0.92\linewidth]{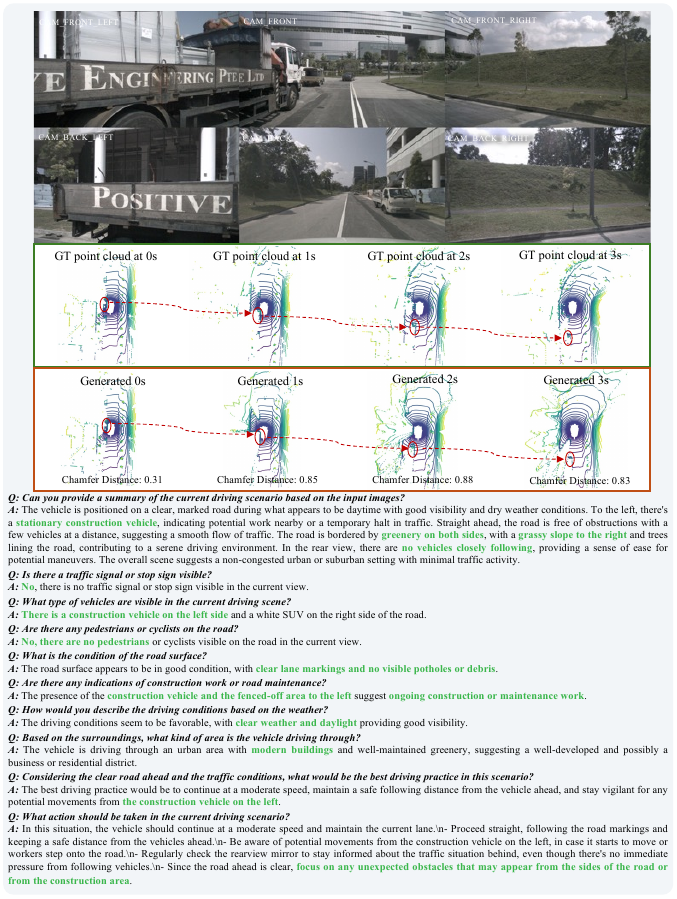}
        \subcaption{}
        \label{fig:vis2}
    \end{subfigure}
\end{figure*}
\begin{figure*}[ht]
\ContinuedFloat
    \begin{subfigure}{\linewidth}
        \centering
        \includegraphics[width=0.93\linewidth]{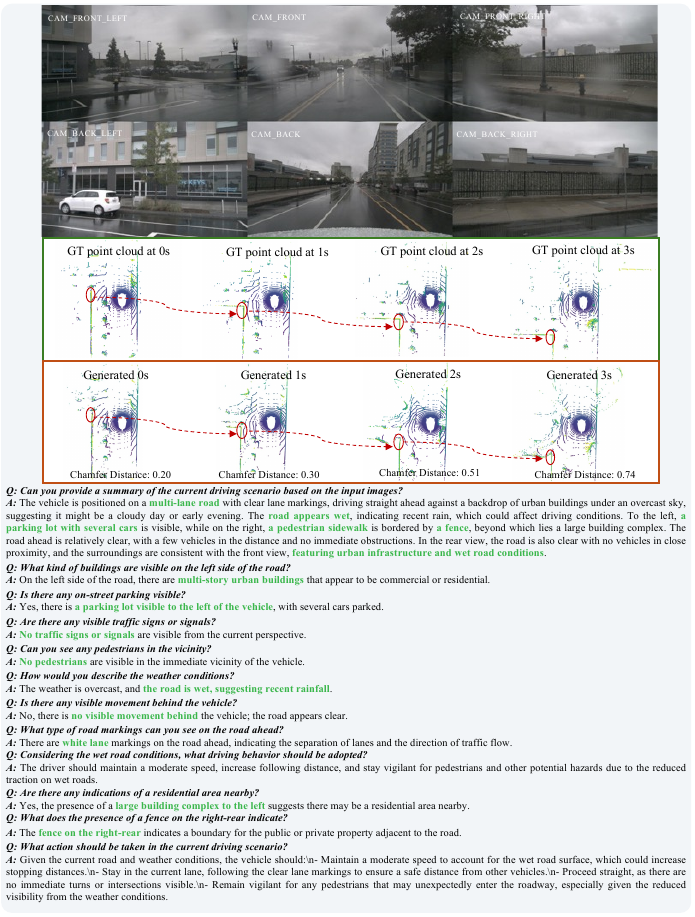}
        \subcaption{}
        \label{fig:vis3}
    \end{subfigure}
\end{figure*}
\begin{figure*}[ht]
\ContinuedFloat
    \begin{subfigure}{\linewidth}
        \centering
        \includegraphics[width=0.925\linewidth]{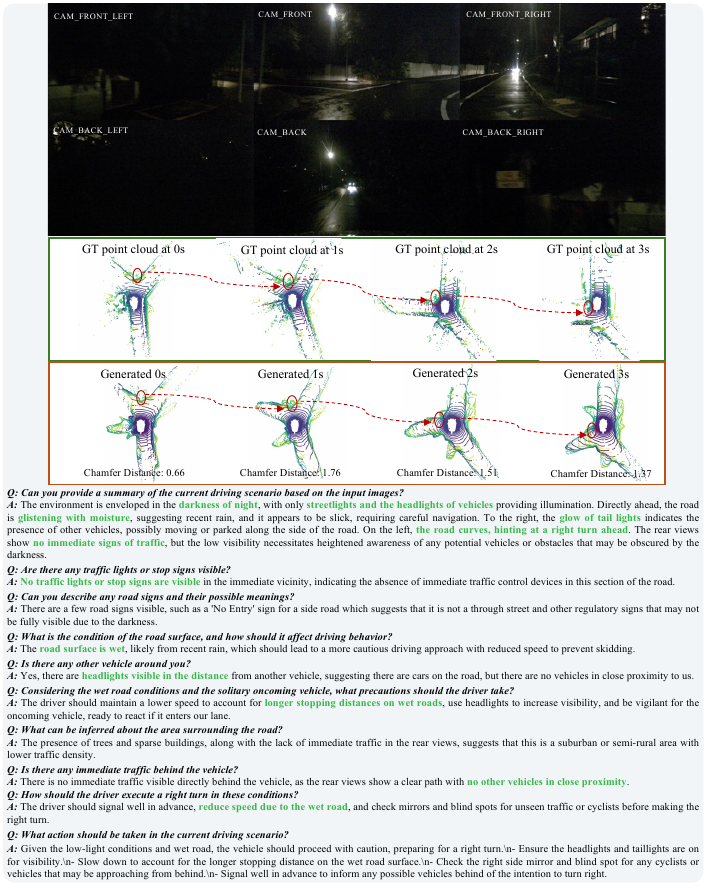}
        \subcaption{}
        \label{fig:vis4}
    \end{subfigure}
    \caption{Qualitative results for future generation and scene understanding. From top to bottom, each sub-figure displays the multi-view input of the current scene, the ground truth scene evolution, the generated scene evolution, and the scene understanding result.}
    \label{fig:vis}
\end{figure*}

\newpage
{
    \small
    \bibliographystyle{ieeenat_fullname}
    \bibliography{main}
}

\end{document}